\documentclass[11pt,a4paper]{article}
\usepackage{times,latexsym}
\usepackage{url}
\usepackage[T1]{fontenc}
\usepackage{tcolorbox}
\usepackage{framed}
\usepackage{booktabs}
\usepackage{dsfont}
\usepackage{CJKutf8}
\usepackage{amsfonts}

\usepackage{amsmath}
\usepackage{amsthm}
\usepackage{mathtools}
\usepackage{enumitem}
\usepackage{tabularx}
\usepackage{multirow}
\usepackage{adjustbox}
\usepackage{subcaption}
\usepackage[utf8]{inputenc}
\usepackage{makecell}
\usepackage{listings}
\usepackage{tcolorbox}  
\usepackage{xcolor}     

\newtcolorbox{promptbox}{
  colback=gray!5,      
  colframe=black!35,   
  boxrule=0.4pt,       
  arc=1mm,             
  left=3pt,right=3pt,  
  top=2pt,bottom=2pt,
  fontupper=\footnotesize\ttfamily 
}

\usepackage{graphicx}
\newtheorem{theorem}{Theorem}
\theoremstyle{theorem*}
\newtheorem*{theorem*}{Theorem}
\newtheorem{proposition}{Proposition}

\theoremstyle{corollary}
\newtheorem{corollary}{Corollary}
\theoremstyle{definition}

\theoremstyle{assumption}

\theoremstyle{remark}

\theoremstyle{observation}

\tcbuselibrary{breakable, skins, theorems}

\usepackage{color}
\usepackage{algorithm}
\usepackage{algorithmic}
\usepackage{newfloat}
\usepackage{listings}
\DeclareCaptionStyle{ruled}{labelfont=normalfont,labelsep=colon,strut=off} 
\floatstyle{ruled}
\newfloat{listing}{tb}{lst}{}
\floatname{listing}{Listing}

\usepackage[acceptedWithA]{tacl2021v1}
\usepackage{cleveref}

\usepackage{xspace,mfirstuc,tabulary}

\newif\iftaclinstructions
\taclinstructionsfalse 
\iftaclinstructions
\renewcommand{\confidential}{}
\renewcommand{\anonsubtext}{(No author info supplied here, for consistency with
TACL-submission anonymization requirements)}
\newcommand{\instr}
\fi

\iftaclpubformat 

\else

\fi

\title{MO-GRPO: Mitigating Reward Hacking of \\
Group Relative Policy Optimization on Multi-Objective Problems}

\author{
Yuki Ichihara$^{1}$ \qquad
Yuu Jinnai$^{2}$ \qquad
Tetsuro Morimura$^{2}$ \qquad
Mitsuki Sakamoto$^{2}$\\
{\bf Ryota Mitsuhashi}$^{2}$ \qquad
{\bf Eiji Uchibe}$^{3}$\\[0.5em]
$^{1}$Nara Institute of Science and Technology \qquad
$^{2}$CyberAgent \qquad\\
$^{3}$Advanced Telecommunications Research Institute International\\
}

\date{}

\begin{document}
\maketitle
\begin{table*}[h]
  \centering
  \begin{tabularx}{\textwidth}{cXcc}
    \toprule

    Instruction & \textbf{Translate the following English into easily readable Japanese.}\textbackslash{}nOver the past decade, our lives have changed through technology, with many working from home, ... & jReadability\,$\uparrow$ & BLEURT\,$\uparrow$ \\
    \midrule
    GRPO  & {\color{red}Over the past ten years, our lives have changed a lot because of technology. ...} & {\color{red} $\mathbf{0.99}$} & 0.57 \\
    \midrule
    MO-GRPO  &
      \makecell[Xt]{\begin{CJK}{UTF8}{ipxm}過去の10年で、技術の進化により、多くの人がホームワークから仕事をしているようになり...
      \end{CJK} \\
      {\color{gray} (Translation: In the past decade, technological advances have enabled many people to work from home work...)}}&
      0.40 & 0.69 \\
    \bottomrule
  \end{tabularx}
  \caption{(Machine translation) Generation examples of GRPO and MO-GRPO by Llama (Llama-3.2-3B-Instruct). 
  \textbf{GRPO optimizes only the Japanese readability score (jReadability) by avoiding using difficult Japanese words, eventually stops using any Japanese characters}, ignoring the translation accuracy score (BLEURT), resulting in generating non-Japanese text, which defeats the purpose of the translation. On the other hand, MO-GRPO evenly optimizes both objectives, achieving improvement on both objectives as intended.
  }  \label{tab:example-Llama}
\end{table*}
\begin{abstract}
Group Relative Policy Optimization (GRPO) has been shown to be an effective algorithm when an accurate reward model is available. However, such a highly reliable reward model is not available in many real-world tasks. In this paper, we particularly focus on multi-objective settings, in which we identify that GRPO is vulnerable to reward hacking, optimizing only one of the objectives at the cost of the others.
To address this issue, we propose MO-GRPO, an extension of GRPO with a simple normalization method to reweight the reward functions automatically according to the variances of their values.
We first show analytically that MO-GRPO ensures that all reward functions contribute evenly to the loss function while preserving the order of preferences, eliminating the need for manual tuning of the reward functions' scales.
Then, we evaluate MO-GRPO experimentally in three domains: (i)  the multi-armed bandits problem, (ii) machine translation tasks on the WMT benchmark (En-Ja, En-Zh), and (iii) the instruction following task.
MO-GRPO achieves stable learning by evenly distributing correlations among the components of rewards, outperforming GRPO, showing MO-GRPO to be a promising algorithm for multi-objective reinforcement learning problems.
\end{abstract}
\section{Introduction}\label{sec:intro}
\textbf{Reward hacking} is a phenomenon in which an agent overfits to a misspecified reward model, failing to optimize for the true intended objective \cite{NEURIPS2022_3d719fee,pmlr-v202-gao23h,rafailov2024scaling}.
Recent work proposes Group Relative Policy Optimization (GRPO; \citealt{shao2024deepseekmath,liu2025understanding}) to enhance the reasoning capability of Large Language Models (LLMs; \citealt{liu2024deepseek,yang2025qwen3}) with high-accuracy reward models; however, it is difficult to obtain them in many real-world tasks.

In this work, we evaluate GRPO's performance in a specific scenario where we only have access to under-specified reward models that do not accurately represent the task's objective on their own.
In particular, we study a practical scenario where we specify the intended behavior of the agent using multiple reward models. 

Our analysis shows that the loss function of the GRPO leads to optimizing the reward functions with higher variances while ignoring the lower ones, which may result in an undesirable policy.
We empirically evaluate GRPO on multi-objective reinforcement learning problems, where we also observe reward hacking behavior of GRPO that it tends to ignore rewards with lower variances and only optimizes the ones with higher variances, resulting in an unintended behavior (e.g., hacking the readability objective while ignoring the translation consistency; Table~\ref{tab:example-Llama}).

To resolve this problem, we propose GRPO with a simple automated normalization method to the objectives, which we call \textbf{MO-GRPO (Multi-Objective Group Relative Policy Optimization)}. MO-GRPO normalizes the advantage functions for each objective so that their variances are scaled evenly.
This ensures that all reward functions contribute equally to updating the policy, regardless of their variance scales (Theorem~\ref{pro:MO-GRPO-std}). 
In this way, it prevents any objectives from being ignored in the training process, mitigating the reward hacking behavior of GRPO on multiple objectives. 
Our normalization technique maintains the original preference ordering under positive affine transformations of reward scales, even in cases where GRPO fails to preserve this ordering.

We evaluate MO-GRPO experimentally on multi-armed bandit, machine translation \cite{kocmi-etal-2024-findings} problems, and instruction following task with multiple objectives. The result shows that MO-GRPO successfully mitigates reward hacking in the problem settings where GRPO incurs reward hacking (Table~\ref{tab:example-Llama}), resulting in a policy that is desirable for the given task.

\section{Group Relative Policy Optimization (GRPO)}

GRPO is a reinforcement learning algorithm \cite{shao2024deepseekmath} which is usually used for online and on-policy learning.
For a given state, a policy generates multiple outputs and learns to generate outputs with higher relative reward scores compared to the rest of the outputs. 
$p_{\mathcal{Q}}$ is the distribution over the initial state (prompt) ($q \sim p_{\mathcal{Q}}$). The policy $\pi_\theta\left(\cdot \mid q\right)$ outputs the action (sentence) $o_g$ based on the initial state $q$ from the action space. Formally, let $R_i$ be the $i$-th reward function, which is a mapping from a prompt-output pair to a scalar value, and assume that there are $K$ reward functions.
For each prompt $q$, GRPO samples a group of outputs $ \mathbf{o} = \left\{o_1, o_2,...,o_G\right\}$ from the old policy $\pi_{\theta_{\text {old}}}$ and then optimizes the policy model by maximizing the following objective:
\begin{align}
\mathcal{J}(\pi_\theta) 
&= \mathbb{E}
\Biggl[
 \sum_{g=1}^G \frac{1}{G} \frac{1}{|o_g|} 
         \frac{\pi_\theta\bigl(o_{g}\mid q\bigr)}%
              {\pi_{\theta_\mathrm{old}}\bigl(o_{g}\mid q\bigr)}
         A_{g}\nonumber
\\&- \beta \text{KL}(\pi_\theta, \pi_{\theta_{\text{ref}}})
\Biggr], \label{eq:grpo_obj}
\end{align}
where $\beta$ is a hyperparameter, and $\text{KL}$ is Kullback–Leibler (KL) divergence. 

$A_g$ represents the normalized advantage value of the sentence $o_g$ using $K$ reward models:
\begin{equation}\label{eq:advantage}
A_{g}=\frac{
     \sum_{i=1}^K R_i(q,o_g
)
     -
     \mathrm{mean}_\mathbf{o}\bigl(\sum_{i=1}^K R_i(q,\mathbf{o})\bigr)}{\mathrm{std}_\mathbf{o}\bigl(\sum_{i=1}^K R_i(q,\mathbf{o})\bigr)}.
\end{equation}

The advantage value $A_{g}$ is computed without normalizing the scale of the reward functions (Eq.~\ref{eq:advantage}). 
Consequently, when rewards differ in scale or variance, the advantage value can be dominated by the value from a high variance reward function.
\begin{theorem}[Correlation between reward function and advantage function with GRPO]\label{theorem:grpo_std_un}
Assume the $G\rightarrow\infty$. The correlation coefficient between an individual reward function $R_i$ and the advantage $A_g$ is the ratio of $R_i$'s standard deviation $\sigma_i$ to the standard deviation of the total reward $\sigma$. 
    \begin{equation}
        \operatorname{Corr}(R_i(q,o_g),A_g) = \frac{\sigma_i^2 +X_i}{\sigma \sigma_i}
    \end{equation}
    where $X_i =\sum_{j \neq i} \operatorname{Cov}(R_i, R_j)$, $\operatorname{Cov}(\cdot,\cdot)$ is covariance.
\end{theorem}
The proof is in Appendix~\ref{appendix:corr}. 
\begin{figure*}[t]
        \centering
        \includegraphics[width=0.8\linewidth]{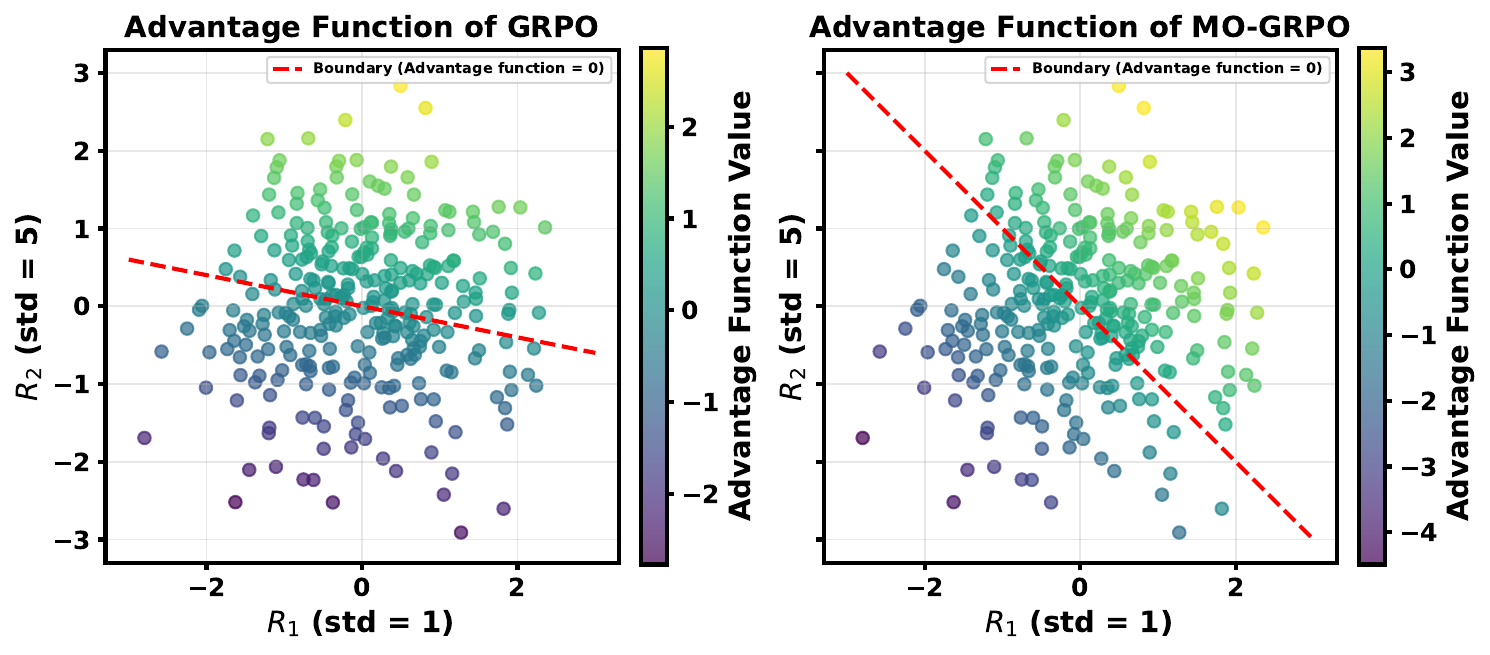}
        \caption{(Simulated experiment) Comparison of the advantage values of GRPO and MO-GRPO on a toy example with two reward functions with different sizes of variances ($1$ and $5$).
        The advantage values of GRPO (left figure) are dominated by the high variation reward ($R_2$), indicating that the algorithm is sensitive to the relative scales of the rewards. In contrast, the advantage values of MO-GRPO (right figure) are invariant with the scale of the reward models, which shows that MO-GRPO is an easy-to-use algorithm for multi-objective learning tasks that does not require manual tuning of the reward models to avoid reward hacking. 
        }
        \label{fig:advanatge}
\end{figure*}
\paragraph{Remark (Finite-$G$ implementation).}
Theorem~\ref{theorem:grpo_std_un} is stated for population quantities, i.e., the advantage is defined using the population mean and standard deviation of the \emph{total reward}. In practice, GRPO computes the advantage using group-wise (empirical) normalization over a finite number of samples $G$:
Accordingly, the population variances/covariances in Theorem~\ref{theorem:grpo_std_un} are replaced by their empirical counterparts computed from the same group:
\begin{align}
\hat\sigma_i^2
&=
\widehat{\mathrm{Var}}(R_i), \qquad
\widehat{\sigma}^2
=
\widehat{\mathrm{Var}}(R_{\mathrm{tot}}), \\
\hat X_i
&=
\sum_{j\neq i}\widehat{\mathrm{Cov}}(R_i,R_j).
\end{align}
where $R_{\mathrm{tot}}(q,o)=\sum_{i=1}^K R_i(q,o)$.
Therefore, for finite $G$, the correlation becomes a random (group-dependent) plug-in estimate
$\widehat{\mathrm{Corr}}(R_i,\hat A)$ rather than a constant.
As $G\to\infty$, $(\hat\sigma_i^2,\widehat{\sigma}^2,\hat X_i)$ converge to $(\sigma_i^2,\sigma^2,X_i)$, and the empirical correlation concentrates around the population expression in Theorem~\ref{theorem:grpo_std_un}.

Theorem~\ref{theorem:grpo_std_un} shows that the advantage function in GRPO is more strongly correlated with reward components that exhibit higher variance.
This shows that GRPO learns to optimize reward functions with higher variances than the lower ones and may lead to unintended behavior, which we show empirically in Section~\ref{sec:exp}.

\section{Multi-Objective GRPO (MO-GRPO)}
To solve the problem that GRPO is affected by the scale of the variance of the reward functions, we propose Multi-Objective GRPO (MO-GRPO). By computing a separate advantage function for each reward, our framework adjusts for differences in reward variance and enables more stable learning.
\begin{align}
\mathcal{J}(\pi_\theta) 
&= \mathbb{E}
\Biggl[
 \sum_{g=1}^G \frac{1}{G} \frac{1}{|o_g|} 
         \frac{\pi_\theta\bigl(o_{g}\mid q\bigr)}%
              {\pi_{\theta_\mathrm{old}}\bigl(o_{g}\mid q\bigr)}
         \color{red}A_g^{\mathrm{MO}}\color{black}\nonumber\\&- \beta \text{KL}(\pi_\theta, \pi_{\theta_{\text{ref}}})
\Biggr], \label{eq:mogrpo_obj}
\end{align}
where $A_g^{\mathrm{MO}}$ is defined as follows:
\begin{equation}\label{eq:mo-advantage}
\color{red}A_g^{\mathrm{MO}}\color{black}=\color{red}\sum^K_{i=1}\color{black}\frac{
    R_i(q, o_g)-
     \mathrm{mean}_\mathbf{o} \bigl(R_i(q, \mathbf{o})\bigr)}{\mathrm{std}_\mathbf{o}\bigl(R_i(q, \mathbf{o})\bigr)}.
\end{equation}

Note that \textbf{MO-GRPO rescales the reward \textbf{\textit{individually}}, then aggregating over the reward functions}, whereas \textbf{vanilla GRPO rescales it after all the reward values are aggregated into a single value (Equation~\ref{eq:advantage})}.
Thus, MO-GRPO ensures a consistent correlation between each advantage function and its corresponding reward function.
\begin{theorem}[Correlation between a reward function and advantage function with MO-GRPO]\label{pro:MO-GRPO-std}
Assume that the number of samples $G \to \infty$. The correlation of the advantage functions $A_g^{\mathrm{MO}}$ with each reward function $R_i$ for any $o_g$ remains constant.
\begin{align}
    \operatorname{Corr}(R_i(q,o_g),A_g^{\mathrm{MO}})&=\frac{1+Z_i}{\sqrt{K+Y}}
\end{align}
where $Y = \sum_{j =1}^K\sum_{l \neq j} \frac{\operatorname{Cov}\left(R_l, R_j\right)}{\sigma_l \sigma_j}$ and $Z_i = \sum_{ j \neq i} \frac{\operatorname{Cov}\left(R_i, R_j\right)}{\sigma_i \sigma_j}$, $\operatorname{Cov}(\cdot,\cdot)$ is covariance. 
\end{theorem}
The proof is in Appendix~\ref{appendix:MO-GRPO-un}.
Theorem~\ref{pro:MO-GRPO-std} assumes $G\rightarrow\infty$, however, MO-GRPO still improves and avoids reward hacking with a small sample size in experiments (Table~\ref{tab:num_size}).

\paragraph{Remark (Finite-$G$ implementation).}
Theorem~\ref{pro:MO-GRPO-std} is stated for the true-normalized advantage, where each reward is centered and scaled by the true mean and standard deviation $(\mu_i,\sigma_i)$.
In practice, MO-GRPO computes the advantage using group-wise (empirical) normalization over a finite number of samples $G$:
Accordingly, the quantities $Y$ and $Z$ in Theorem~\ref{pro:MO-GRPO-std} are also replaced by their empirical counterparts computed from the same group:
\begin{align}
\hat Y&=\sum_{j=1}^K\sum_{l\neq j}\frac{\widehat{\mathrm{Cov}}(R_l,R_j)}{\hat\sigma_l\hat\sigma_j},
\\
\hat Z_i&=\sum_{j\neq i}\frac{\widehat{\mathrm{Cov}}(R_i,R_j)}{\hat\sigma_i\hat\sigma_j},
\end{align}
where $\widehat{\mathrm{Cov}}(\cdot,\cdot)$ denotes the sample covariance over the $G$ candidates.
Therefore, for finite $G$, the correlation becomes a random (group-dependent) plug-in estimate
$\widehat{\mathrm{Corr}}(R_i,\hat A^{\mathrm{MO}})$ rather than a constant.
As $G\to\infty$, $(\hat\sigma_i,\widehat{\mathrm{Cov}})$ converge to $(\sigma_i,\mathrm{Cov})$, and the empirical correlation concentrates around the population expression in Theorem~\ref{pro:MO-GRPO-std}.

Theorem~\ref{pro:MO-GRPO-std} shows that the advantage function in MO-GRPO is roughly equal to $\frac{1}{\sqrt{K}}$ for every reward function $R_i$, with some effect of correlation between the reward functions.
Specifically, if all the reward functions are uncorrelated with each other, then the correlation of the reward and the advantage is exactly $\frac{1}{\sqrt{K}}$ for all the reward functions:
\begin{corollary}[Correlation between a reward function and advantage function with MO-GRPO under certain assumptions]\label{coro:MO-GRPO-std}
Assume that the $K$ reward functions $R_i$ are mutually uncorrelated and the number of samples $G \to \infty$. The correlation of the advantage functions $A_g^{\mathrm{MO}}$ with each reward function $R_i$ for any $o_g$ remains constant.
    \begin{equation}
        \operatorname{Corr}(R_i(q,o_g), A_g^{\mathrm{MO}})  
        =\frac{1}{\sqrt{K}}
    \end{equation}
\end{corollary}
The proof follows immediately from Theorem ~\ref{pro:MO-GRPO-std}.
The result shows that the reward functions of the MO-GRPO have roughly the same amount of influence on the policy update regardless of their variances.
Thus, MO-GRPO does not ignore reward functions with lower variances, which could lead to unintended behavior.
We show this property empirically in Section~\ref{sec:exp}.

\paragraph{Simulated experiment.}
Figure~\ref{fig:advanatge} shows the comparison of the advantage functions of GRPO (Eq.~\ref{eq:advantage}) and MO-GRPO (Eq.~\ref{eq:mo-advantage}) with two reward functions with low and high variances.
The reward functions return a value sampled from a normal distribution $R_1 \sim \mathcal{N}\left(1,1^2\right)$ and $R_2 \sim \mathcal{N}\left(1,5^2\right)$.
The advantage value by GRPO is significantly influenced by $R_2$ while the effect of $R_1$ is negligible. This motivates the agent to maximize the value of $R_2$ even at the cost of losing $R_1$. 
Conversely, the advantage value calculated by MO-GRPO successfully considers both reward functions, even when their variances differ significantly.


\paragraph{Invariance to positive affine transformation.}
An additional advantage of MO-GRPO is its invariance under positive affine transformations of reward functions.
\begin{proposition}[Affine Invariance of MO-GRPO Advantage]\label{theorem:scale}
Let $o_a$ and $o_b$ be two possible outputs, and let $\mathcal{R} = \{R_i\}_{i=1}^K$ be a set of reward functions. Consider a transformed set $\mathcal{R}' = \{R'_i = a_i R_i + b_i\}_{i=1}^K$ with $a_i > 0$.
Then, the preference ordering induced by MO-GRPO is invariant under such positive affine transformations:
\begin{align*}
A^{\mathrm{MO}}_a \ge A^{\mathrm{MO}}_b\iff
A^{\mathrm{MO'}}_a\ge A^{\mathrm{MO'}}_b
\end{align*}
\end{proposition}
where $A^{\mathrm{MO'}}_a = \sum^K_{i=1}\color{black}\frac{
    R'_i(q, o_a)
     \;-\;
     \mathrm{mean}_\mathbf{o} \bigl(R'_i(q, \mathbf{o})\bigr)}{\mathrm{std}_\mathbf{o} \bigl(R'_i(q, \mathbf{o})\bigr)}$.
     
The proof of Proposition~\ref{theorem:scale} is in Appendix~\ref{appendix:scale}.
Meanwhile, Theorem~\ref{pro:MO-GRPO-std} shows that MO-GRPO does not require engineering efforts to normalize the scale of the reward functions \textit{relative} to other reward functions, Proposition~\ref{theorem:scale} shows that it does not need to normalize the \textit{absolute} scale of the reward functions.
This makes MO-GRPO a practically useful algorithm for real-world problems where the scale of the reward models is unclear (e.g., neural models) or instance-dependent.
Conversely, this property does not hold with GRPO.
\begin{proposition}\label{prop:scale}
    The preference ordering induced by GRPO (and Dr .GRPO) is not invariant under positive affine transformations.
\end{proposition}
The proof is in Appendix~\ref{appendix:prop_scale}.

\paragraph{Summary.}
Unlike GRPO (Theorem~\ref{theorem:grpo_std_un}), the advantage function of MO-GRPO is built keeping the correlation with each reward function constant (Theorem~\ref{pro:MO-GRPO-std}). In addition, it is invariant with the rescaling of the reward functions with positive affine transformation (Proposition~\ref{theorem:scale}).

These properties make MO-GRPO an easy-to-use algorithm. It can use off-the-shelf reward functions without requiring manual tuning of the reward values to fit them to target tasks.

\begin{figure}[t]
    \centering
    \includegraphics[width=0.9\linewidth]{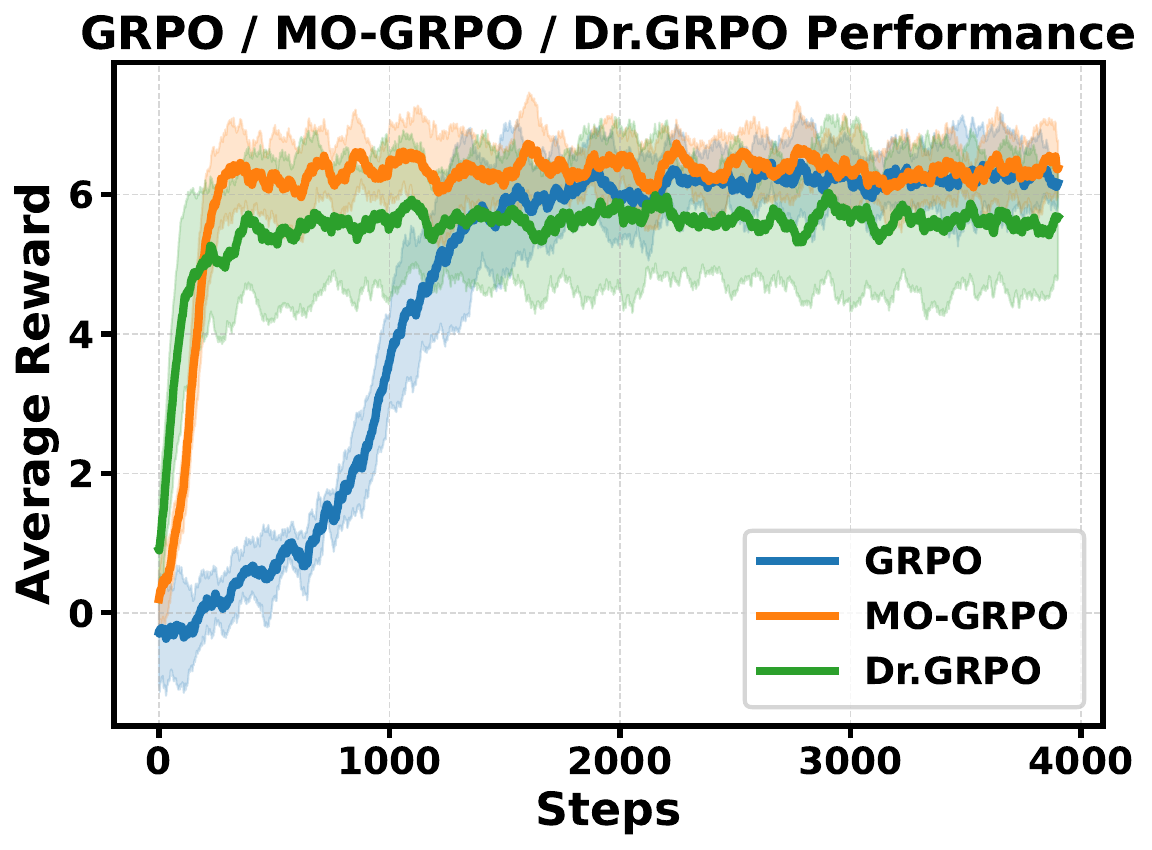}
    \caption{(Multi-armed bandit) This figure illustrates the average rewards obtained by the sum of the three reward functions: GRPO, MO-GRPO, and Dr. GRPO. As the figure shows, MO-GRPO finds a better policy faster than GRPO and Dr. GRPO.}
    \label{fig:ba}
\end{figure}

\begin{figure}[htb!]
    \centering
    \begin{subfigure}[c]{0.8\columnwidth}
    \includegraphics[width=\linewidth]{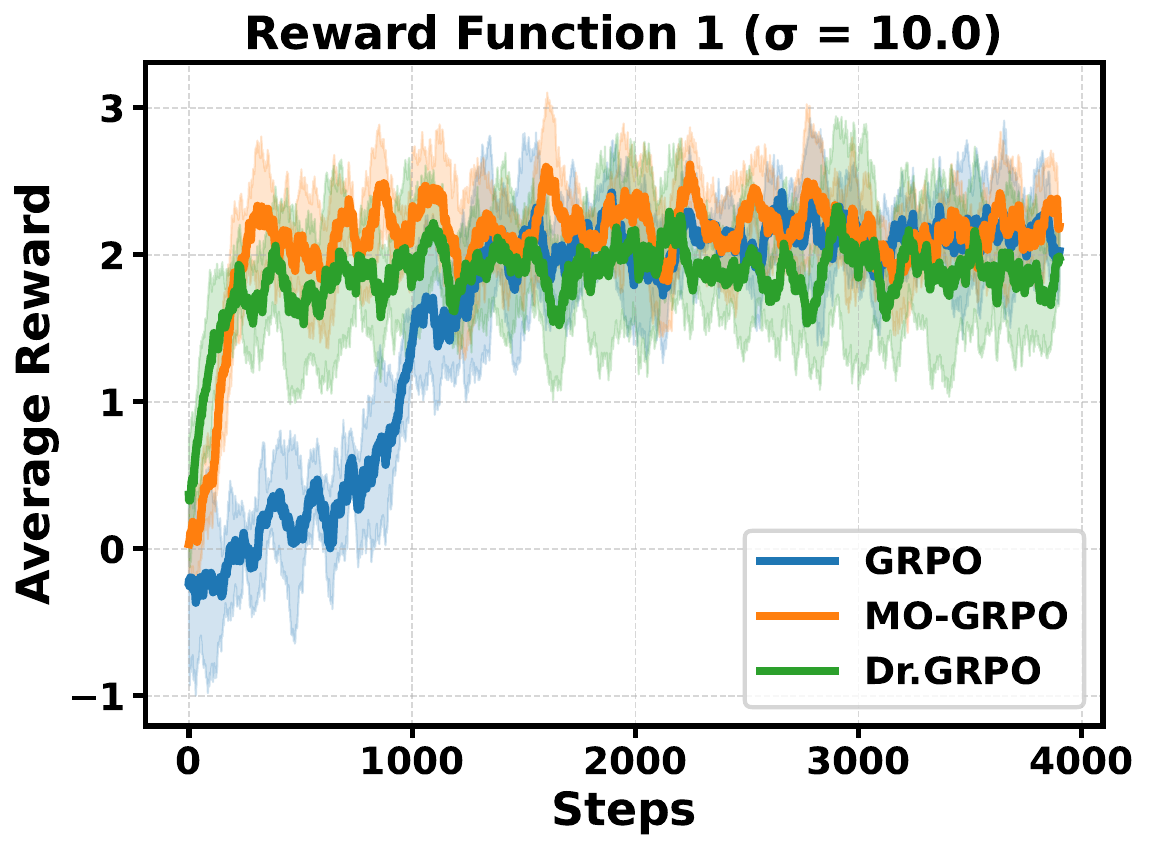}
    \end{subfigure}
    \begin{subfigure}[c]{0.8\columnwidth}
    \includegraphics[width=\linewidth]{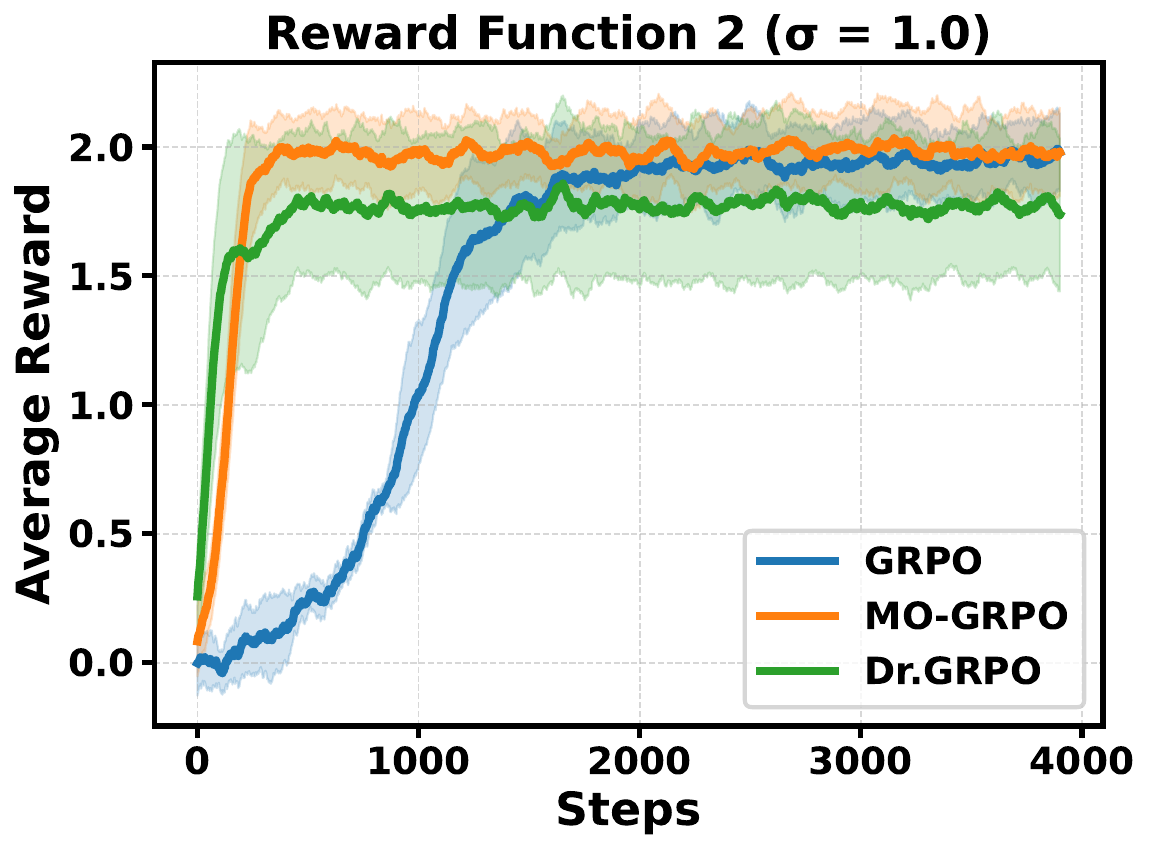}
    \end{subfigure}
    \begin{subfigure}[c]{0.8\columnwidth}
    \includegraphics[width=\linewidth]{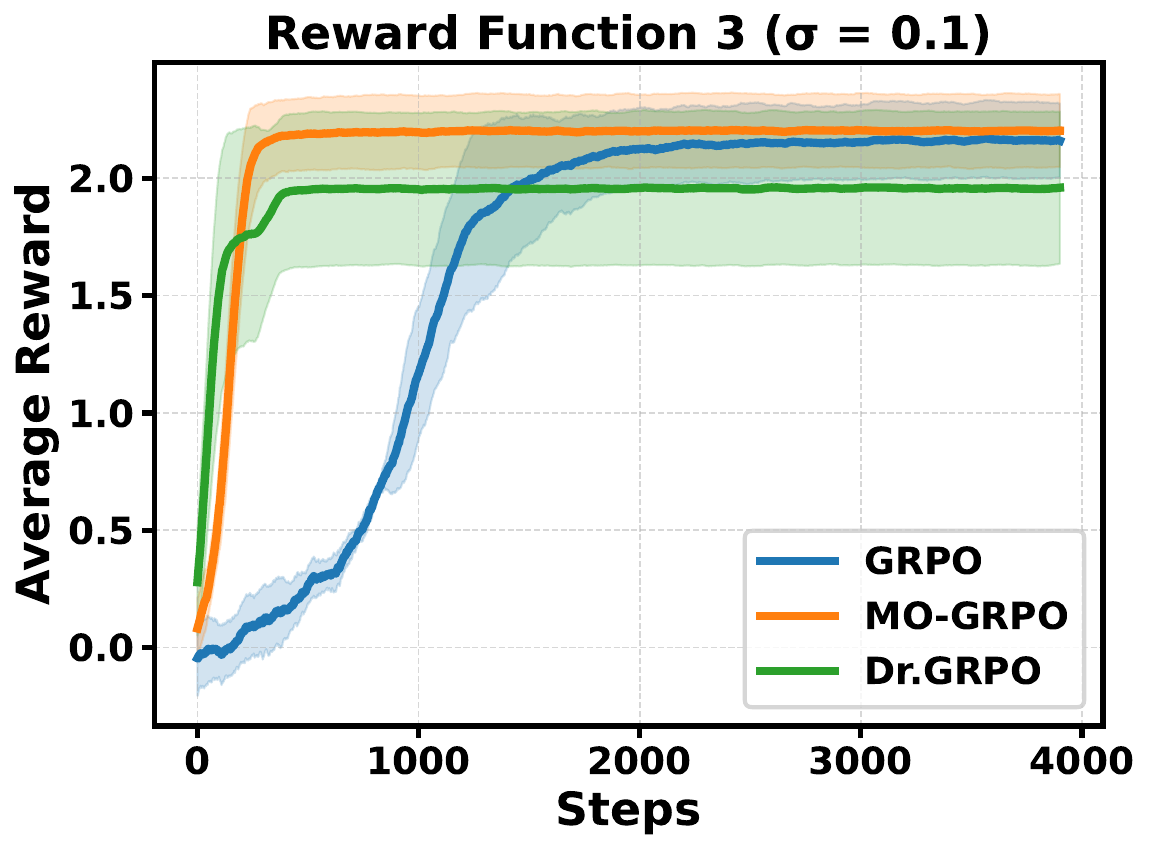}
    \end{subfigure}
    \caption{(Multi-arm bandit) Comparison of the three reward functions with varying variances ($10$, $1$, and $0.1$) obtained by GRPO, Dr. GRPO, and MO-GRPO. While GRPO and Dr. GRPO fail or are slow to learn the reward functions with lower variances ($R_2$ and $R_3$), MO-GRPO successfully optimizes all three reward functions regardless of the scale of the variances.}
    \label{fig:ba-breakdown}
\end{figure}
\section{Experiment}\label{sec:exp}
We conduct experiments on three tasks: (1) multi-armed bandit, (2) machine translation, and (3) instruction following task. 
We compare three methods: GRPO, Dr. GRPO, and MO-GRPO\footnote{The simplest implementation example of MO-GRPO is shown in Appendix~\ref{appendix:mo-grpo}.}.
\begin{figure*}[t]
  \centering

  \includegraphics[width=0.9\linewidth]{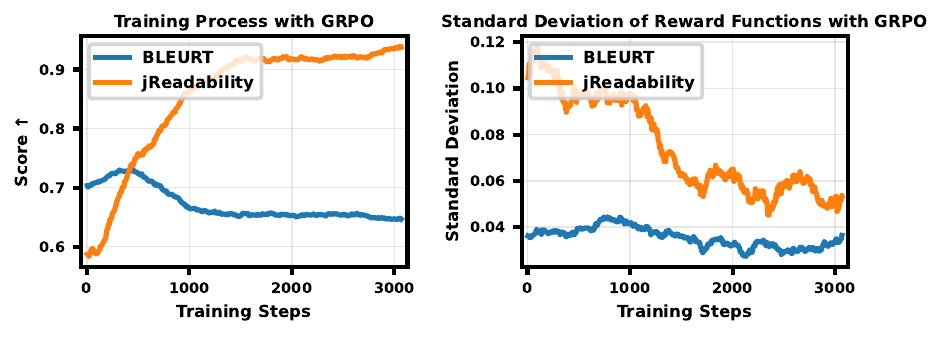}


  \includegraphics[width=0.60\linewidth]{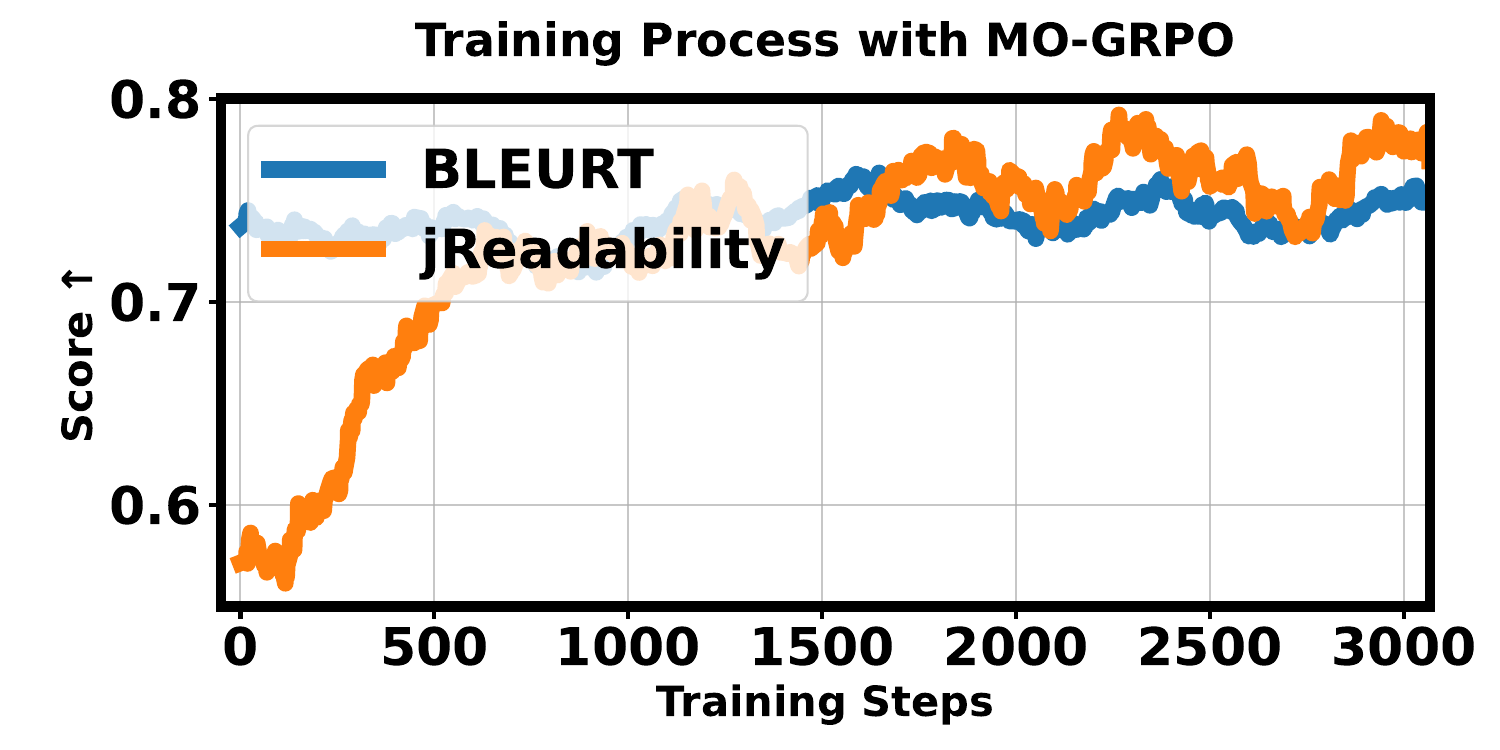}

  \caption{(Machine translation) Training curves on WMT En--Ja with BLEURT and jReadability rewards (Sarashina; sarashina2.2-3b-instruct-v0.1). 
  \textbf{Top:} GRPO overfits jReadability at the expense of BLEURT; the standard deviation of jReadability is consistently larger than that of BLEURT. 
  \textbf{Bottom:} MO-GRPO mitigates this issue, avoiding overfitting to jReadability and preventing BLEURT deterioration}
  \label{fig:std_mo_combined}
\end{figure*}
\subsection{Multi-Armed Bandit}
We first conduct experiments on a simple multi-armed bandit environment to observe the behavior of GRPO and MO-GRPO in a controlled environment.
We set the number of arms (actions) $k$ to $50$, and there are three stochastic reward functions $R_1, R_2$, and $R_3$. The episode length is fixed to $5000$ steps.
The expected return of the arm $\mu_k$ is chosen at random from a normal distribution of $\mathcal{N}(0, 1)$ at the beginning of the episode and is fixed throughout the episode.
The three reward functions output the reward value of $\mu_k$ plus additional stochastic noise to it as follows:
    $R_1(k) \sim \mathcal{N}(\mu_k, 10^2), 
    R_2(k) \sim \mathcal{N}(\mu_k, 1^2) - 0.1 R_1(k)$, and $R_3(k) \sim \mathcal{N}(\mu_k, 0.1^2)$.
$R_1(k)$ outputs a value sampled from a normal distribution of mean $\mu_k$ and standard deviation of $10$. $R_2(k)$ is a sum of a value sampled from a normal distribution of mean $\mu_k$ and standard deviation of $1$, minus the value of $0.1 \times R_1$. Thus, $R_2$ is negatively correlated with $R_1$, making it harder to learn. $R_3$ is a value sampled from a normal distribution of mean $\mu$ and standard deviation of $0.1$.
These reward functions are designed to create a challenging optimization landscape where a high-variance reward function ($R_1$) could dominate the learning signal, potentially overshadowing the gradients from lower-variance and/or negatively correlated reward functions ($R_2$ and $R_3$).

We set the number of actions GRPO, Dr. GRPO, and MO-GRPO samples $G$ to $8$, and we use a neural network with $3$ hidden layers for policy. We conduct experiments with five different random seeds.

Figure~\ref{fig:ba} shows the comparison of the three algorithms on the sum of the three reward functions. The result shows that MO-GRPO achieves a better policy faster than the others. Figure~\ref{fig:ba-breakdown} shows the breakdown of the three reward functions, showing that GRPO and Dr. GRPO fail to learn low variance reward functions ($R_2$ and $R_3$), resulting in suboptimal policies.
The experimental result on a multi-armed bandit problem shows that MO-GRPO is a promising approach for tasks where the reward functions have different scales of variance.

\begin{table*}[t]
  \centering
  \small
  \begin{tabular}{llccc}
    \toprule
     & & \multicolumn{3}{c}{\textbf{WMT24 (En–Ja)}}\\
    \cmidrule(lr){3-5}
    \textbf{Base Model} & \textbf{Method}
      & \textbf{BLEURT$^{\dagger}$}\,$\uparrow$  
      & \textbf{jReadability$^{\dagger}$}\,$\uparrow$ 
      & \textbf{GPT‑Eval$^{\star}$}\,$\uparrow$\\
    \midrule
    \multirow{4}{*}{Sarashina}
      & Base Model          & 0.66 & 0.70 & 50.0\% \\
      & GRPO          & 0.62 & \textbf{0.86} & 66.0\% \\
      & Dr. GRPO       & 0.67 & 0.73 & \underline{68.4\%} \\
      & MO-GRPO (ours) & \textbf{0.69} & 0.76  & \textbf{76.8\%}\\
    \midrule
    \multirow{4}{*}{Qwen}
      & Base Model         & \textbf{0.67} & 0.66 & 50.0\% \\
      & GRPO          & 0.65 & \textbf{0.74} & 53.0\% \\
      & Dr. GRPO       & \textbf{0.67} & 0.66 & \underline{69.6\%}\\
      & MO-GRPO (ours) & \textbf{0.67} & 0.67  & \textbf{88.8\%}\\
    \midrule
    \multirow{4}{*}{Llama}
      & Base Model        & 0.65 & 0.67 & \underline{50.0\%} \\
      & GRPO          & 0.60 & \textbf{0.90} & 35.6\% \\
      & Dr. GRPO       & 0.63 & 0.77 & 42.6\% \\
      & MO-GRPO (ours) & \textbf{0.66} & 0.69 & \textbf{68.8\%} \\
    \midrule
     & & \multicolumn{3}{c}{\textbf{WMT24 (En–Zh)}}\\
    \cmidrule(lr){3-5}
    \textbf{Base Model} & \textbf{Method}
      & \textbf{BLEURT$^{\dagger}$}\,$\uparrow$  
      & \textbf{TRank$^{\dagger}$}\,$\downarrow$ 
      & \textbf{GPT‑Eval$^{\star}$}\,$\uparrow$\\
    \midrule
    \multirow{4}{*}{Qwen}
      & Base Model         & \textbf{0.73} & -2.73 & 50.0\% \\
      & GRPO          & 0.71 & \textbf{-3.20} & 53.9\% \\
      & Dr. GRPO       & \textbf{0.73} & -2.98 & \underline{62.4\%}\\
      & MO-GRPO (ours) & \textbf{0.73} & -2.85  & \textbf{68.1\%}\\
    \midrule
    \multirow{4}{*}{Llama}
      & Base Model        & 0.69 & -2.44 & \underline{50.0\%} \\
      & GRPO          & 0.62 & \textbf{-2.98} & 33.2\% \\
      & Dr. GRPO       & 0.60 & -3.15 & 28.3\% \\
      & MO-GRPO (ours) & \textbf{0.71} & -2.55 & \textbf{71.7\%} \\
    \bottomrule
  \end{tabular}
  \caption{(Machine translation) Translation quality on WMT24 (higher is better). 
  $^{\dagger}$\textbf{BLEURT}, \textbf{jReadability}, and \textbf{TRank} are training objectives and thus susceptible to over‑fitting.
  $^{\star}$\textbf{GPT‑Eval} (against Base Model) is not optimized during training; we therefore regard it as the \emph{primary metric}.  
  Across all three base models, our MO-GRPO improves GPT‑Eval while avoiding excessive optimization of the training‑objective metrics.}
  \label{tab:wmt24-combined}
\end{table*}
\subsection{Machine Translation}\label{sec:WMT} 
We evaluate the performance of MO-GRPO  on machine translation with two objective functions, translation accuracy and readability.
Readability is one of the important objectives in real-world text generation tasks \cite{hasebe2015introducing,trokhymovych-etal-2024-open}. 
It measures the accessibility of the text for a diverse audience, including children and non-native speakers, and is critical for communicating vital information during emergencies such as natural disasters. 

We conduct experiments on English to Japanese (En-Ja) and English to Chinese (En-Zh).
We use the WMT-21, WMT-22, and WMT-23 datasets for training \citep{akhbardeh-etal-2021-findings,freitag-etal-2022-results,freitag-etal-2023-results}, and evaluate on the WMT-24 test set \citep{kocmi-etal-2024-findings}.  

First, we perform the En-Ja translation task in WMT datasets using Sarashina (\texttt{sarashina2.2-3b-instruct-v0.1}), Qwen (\texttt{Qwen2.5-3B-Instruct}) \cite{yang2025qwen3}, and Llama (\texttt{Llama-3.2-3B-Instruct}) \cite{grattafiori2024llama} as the base models. 
For the reward (objective) functions, we adopt (i) BLEURT~\citep{sellam-etal-2020-bleurt} and (ii) jReadability~\citep{hasebe2015introducing} to measure readability in Japanese. 
To evaluate the overall generation quality, we use LLM-as-a-Judge \cite{NEURIPS2023_91f18a12} with GPT-4o-mini (GPT-Eval) so that both the translation accuracy and readability are considered.

Table~\ref{tab:wmt24-combined} shows that, compared to the base model score, GRPO achieved a high jReadability score but at the cost of degrading the BLEURT score. This result leads to the worst win rate score against the base model in three methods. 
In contrast, MO-GRPO almost successfully improved both metrics compared to the base model's score, achieving in BLEURT and jReadability scores, preventing overfitting to jReadability, and MO-GRPO also achieves the highest win rate score.
For supplementary, Dr. GRPO also shows higher values for both metrics compared to the base model, but not as high as MO-GRPO with respect to GPT-Eval.\footnote{Dr. GRPO in this experiment is implemented using trl=0.16.1. This version of Dr. GRPO has no exclusions regarding sentence length normalization, only in the form of removing the standard deviation of the advantage function.}

In detail, MO-GRPO with Sarashina achieves BLEURT score of 0.69. For comparison, when Sarashina is trained with GRPO solely on BLEURT, the score reached 0.70. This close score suggests that MO-GRPO effectively learns to optimize BLEURT without sacrificing other objectives.
Furthermore, the training process of MO-GRPO with Sarashina is shown in Figure~\ref{fig:std_mo_combined} (Bottom), which suggests MO-GRPO avoids overfitting of jReadability and prevents degradation of BLEURT, unlike GRPO (Figure~\ref{fig:std_mo_combined} (Top)). 

Next, we examine the details of the results from other language models such as Qwen and Llama.
The relatively limited improvement of MO-GRPO in Qwen's experiments is likely attributable to Qwen not being a language model specialized for Japanese.
However, Qwen outputs are also confirmed reward hacking behavior from GRPO (such behavior is also not observed with MO-GRPO). 
Llama outputs (Table~\ref{tab:example-Llama}) show that GRPO engaged in reward hacking by outputting English text instead of a Japanese translation, thereby improving the jReadability. This phenomenon again did not occur with MO-GRPO. 

Second, we conduct En-Zh translation task in WMT datasets using Qwen and Llama as base models. For the reward functions, we adopt (i) BLEURT and (ii) TRank~\citep{trokhymovych-etal-2024-open}, which can evaluate the readability of text across multiple languages, including Chinese. Since TRank scores are higher for more difficult texts, multiply the score by $-1$ in this experiment setting during the training (Table~\ref{tab:wmt24-combined} shows the true TRank score, i.e., the score obtained without applying the $-1$ multiplication during evaluation.).

As shown in Table~\ref{tab:wmt24-combined}, MO-GRPO also appropriately treats the two reward functions across Qwen and Llama, similar to the En-Ja task.  Therefore, it consistently achieves higher GPT-Eval win rates than GRPO and Dr. GRPO. GPT-Eval indicates that GRPO and Dr. GRPO are improving with Qwen but exhibit clear reward hacking with Llama. Both methods overfit to TRank at the expense of BLEURT. Interestingly, the trained models output non-Chinese text even though the task is Chinese translation (similarly to what we observed in En-Ja translation task in Table~\ref{tab:example-Llama}).

Table~\ref{tab:error_rate} further quantifies this phenomenon using langdetect, a tool to predict the language of the given text.\footnote{https://pypi.org/project/langdetect/} We find that GRPO and Dr. GRPO frequently produce non-Chinese generations, exploiting TRank, resulting in significantly lower GPT-Eval scores than the base model.

\begin{table}[]
    \centering
     \adjustbox{max width=\columnwidth}{
    \begin{tabular}{lcc}
    \toprule
    \textbf{Method} 
        & \makecell[c]{\textbf{Non-Chinese} (\%)\\\textbf{(w/o Penalty)\,$\downarrow$}} 
        & \makecell[c]{\textbf{Non-Chinese} (\%)\\\textbf{(w/ Penalty)\,$\downarrow$}} \\
    \midrule
    Base Model (Llama) & 14.7\% & 14.7\% \\
    GRPO               & 68.7\% &  1.2\% \\
    Dr. GRPO           & 70.7\% &  1.2\% \\
    MO-GRPO            &  5.6\% &  \textbf{0.6\%} \\
    \bottomrule
    \end{tabular}}
    \caption{The probability of non-Chinese outputs (\textbf{Non-Chinese}) in machine translation.
    The reference set contains 851 Chinese sentences (by langdetect).
    \textbf{Non-Chinese} $=1-\#\text{Chinese}/851$.
    \textbf{w/o Penalty}: TRank only. 
    \textbf{w/ Penalty}: TRank with penalty for non-Chinese outputs.
    MO-GRPO consistently maintains proper outputs under both settings.}
    \label{tab:error_rate}
\end{table}

\paragraph{Ablation study.}
Instead of using MO-GRPO, one may solve the reward hacking by patching the reward function so that it cannot be hacked.
We improve the TRank reward function by giving a huge penalty (score=10) if the generated output is identified as non-Chinese by langdetect. In this way, we can prevent the model to learn to generate non-Chinese texts.
Table~\ref{tab:error_rate} shows that adding a penalty to TRank reduces the probability of non-Chinese outputs (\textbf{Non-Chinese}) for all methods; MO-GRPO still has the lowest probability.
Additionally, Table~\ref{tab:with-langdetect} shows the TRank with penalties under the same experimental settings as Table~\ref{tab:wmt24-combined}. This shows that MO-GRPO outperforms other methods in terms of GPT-Eval, and in this setting as well, GRPO and Dr. GRPO are trained with a focus on TRank penalties, which fluctuate more than BLEURT.
This shows that MO-GRPO is on par with GRPO even if the reward functions are reasonably designed (e.g., problem settings in which GRPO achieves improvement).

\begin{table}[t]
    \centering
    \adjustbox{max width=\columnwidth}{
    \begin{tabular}{lccc}
    \toprule
    \textbf{Method} & \textbf{BLEURT}\,$\uparrow$ & \textbf{TRank w/ Penalty}\,$\downarrow$ & \textbf{GPT-Eval}\,$\uparrow$\\
    \midrule
    Base Model&0.69 & 1.39 & 50\% \\
         GRPO&0.70 & -0.66 & 71.5\% \\
         Dr. GRPO&0.70 & -0.64 & 69.6\%\\
         MO-GRPO&0.71 & -0.47 & \textbf{74.0\%} \\
         \bottomrule
    \end{tabular}}
    \caption{(Machine translation) Translation quality on WMT24 En-Zh (higher is better) with Llama. 
  \textbf{TRank w/ Penalty} penalty non-Chinese outputs.
  MO-GRPO improves GPT‑Eval while avoiding excessive optimization of the training‑objective metrics.}
    \label{tab:with-langdetect}
\end{table}

\subsection{Instruction Following Task}
In this section, we conduct an experiment in AlpacaFarm (training dataset: \texttt{tatsu-lab/alpaca}, eval dataset: \texttt{tatsu-lab/alpaca\_eval}) \cite{NEURIPS2023_5fc47800} to evaluate the performance of MO-GRPO for the generic instruction following task using Qwen and Llama as base models. 
We use RM-Mistral-7B \cite{dong2023raft} and the Length reward function. The Length reward function ($R_{\text{Len}}$) gives a higher reward on the outputs closer to the length of the reference text so that it mitigates the length bias problem \cite{shen-etal-2023-loose,singhal2024a}. 
Length reward function is defined as follows:
\[
R_{\text{Len}} =
\begin{aligned}
&\begin{cases}
\dfrac{L}{0.9L_{\text{ref}}}, & L < 0.9L_{\text{ref}}, \\
1, & 0.9L_{\text{ref}} \leq L \leq 1.1L_{\text{ref}}, \\
\dfrac{1.1L_{\text{ref}}}{L}, & L > 1.1L_{\text{ref}}.
\end{cases}
\end{aligned}
\]
where $L_{\text{ref}}$ is the reference text length, $L$ is the output text length.
Given that RM-Mistral-7B tends to prefer longer outputs \cite{shen-etal-2023-loose,singhal2024a}, the Length reward function is an adversarial objective to it, making the optimization more challenging.

\begin{table}[]
  \centering
  \adjustbox{max width=\columnwidth}{
  
  \begin{tabular}{llcc}
    \toprule
     & & \multicolumn{2}{c}{\textbf{AlpacaFarm}}\\
    \cmidrule(lr){3-4}
    \textbf{Base Model} & \textbf{Method}
      & \textbf{RM-Mistral-7B}\,$\uparrow$  
      & \textbf{Length}\,$\uparrow$ \\
    \midrule
    \multirow{4}{*}{Qwen}
      & Base Model         & 5.55 & 0.42\\
      & GRPO          & 5.81 & 0.36  \\
      & Dr. GRPO       & 6.24 & 0.34 \\
      & MO-GRPO (ours) & 5.51 & 0.44  \\
    \midrule
    \multirow{4}{*}{Llama}
      & Base Model        & 5.26 & 0.42  \\
      & GRPO          & 5.56 & 0.37\\
      & Dr. GRPO       & 5.90 & 0.34 \\
      & MO-GRPO (ours) & 5.28& 0.42\\
    \bottomrule
  \end{tabular}}
  \caption{(AlpacaFarm) Since RM-Mistral and Length have conflicting objectives, the correct answer here is to prevent it from being derived from the base model. GRPO and Dr. GRPO have learned to prioritize RM-Mistral, resulting in a significant sacrifice of Length, but MO-GRPO retains both reward functions almost entirely.}
  \label{tab:alpaca}
\end{table}

Table~\ref{tab:alpaca} shows that GRPO and Dr. GRPO optimize RM-Mistral while decreasing the Length of both Llama and Qwen. In contrast, MO-GRPO attempts to maintain the values of both rewards.
In such adversarial cases where both reward functions are important, the optimal behavior is to remain close to the base model, such as MO-GRPO.

\subsection{Supplementary Results}
\paragraph{Effect of group size.}
To quantify the sensitivity of the performance of MO-GRPO, we vary the number of samples $G\in\{2,4,6\}$ for GRPO and MO-GRPO in Table~\ref{tab:num_size} on En-Ja task.
As shown in Table~\ref{tab:num_size}, GRPO exhibits a trade-off as $G$ increases: jReadability improves (0.77$\rightarrow$0.87) while BLEURT degrades (0.67$\rightarrow$0.64), consistent with over-optimization to the readability reward.
MO-GRPO avoids this reward hacking: it preserves BLEURT (0.68$\rightarrow$0.69) and keeps jReadability stable (0.75$\rightarrow$0.76) across $G$, showing that MO-GRPO is robust to the choice of $G \in \{2, 4, 6\}$.
\begin{table}[t]
  \centering
  \adjustbox{max width=\columnwidth}{
  \begin{tabular}{lccc}
    \toprule
    \textbf{Method} & \textbf{\#Samples} & {\textbf{BLEURT}$\uparrow$} & {\textbf{jReadability}$\uparrow$} \\
    \midrule
    Base Model (Sarashina) & -- & 0.66 & 0.70 \\
    \midrule
    GRPO    & 2 & 0.67 & 0.77 \\
    MO-GRPO & 2 & 0.68 & 0.75 \\
    \midrule
    GRPO    & 4 & 0.65 & 0.83 \\
    MO-GRPO & 4 & 0.69 & 0.75 \\
    \midrule
    GRPO    & 6 & 0.64 & 0.87 \\
    MO-GRPO & 6 & 0.69 & 0.76 \\
    \bottomrule
  \end{tabular}}
  \caption{Performance of GRPO and MO-GRPO with varying numbers of samples on BLEURT and jReadability.}
  \label{tab:num_size}
\end{table}

\begin{figure}
    \centering
    \includegraphics[width=\linewidth]{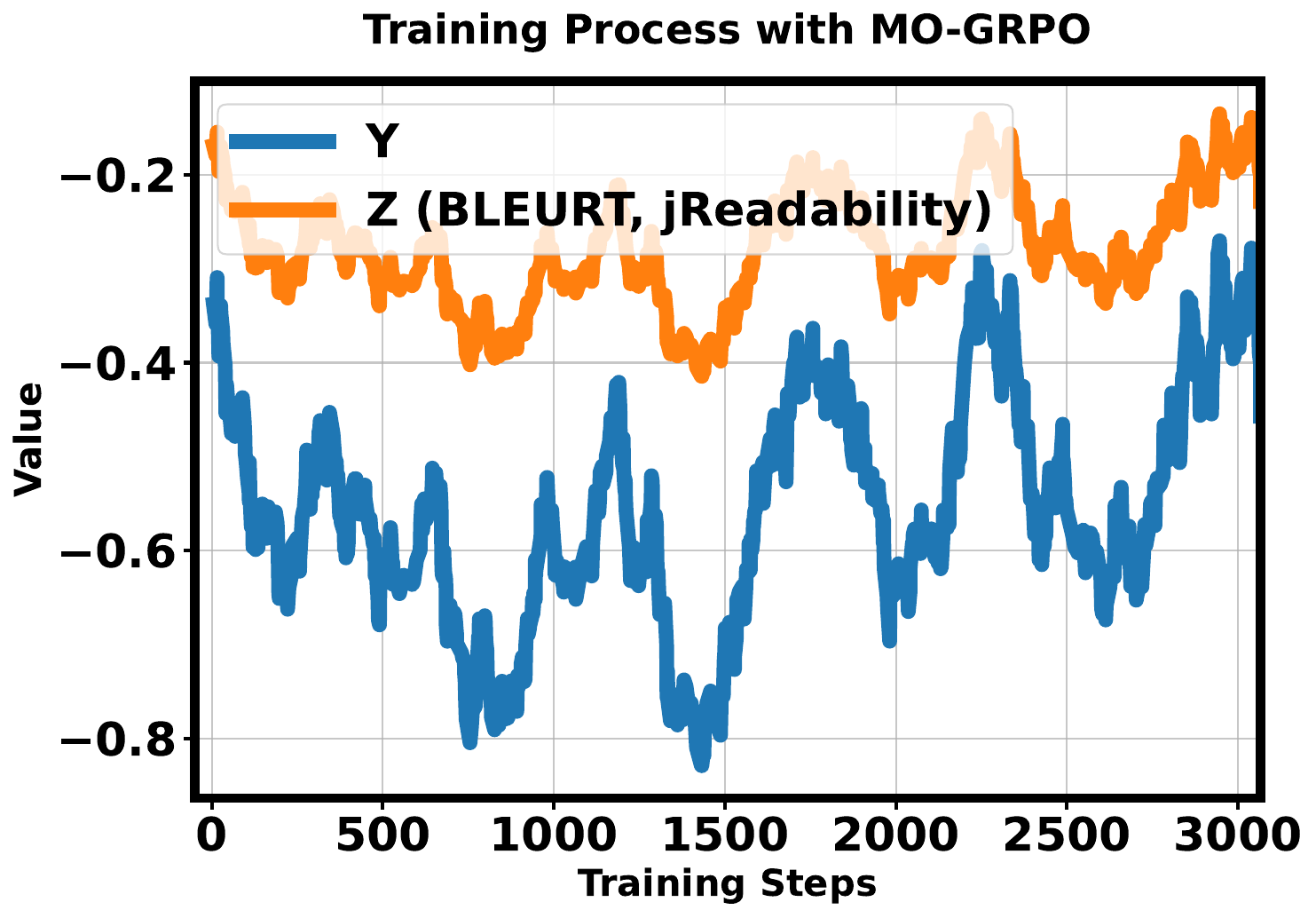}
    \caption{Empirical correlation statistics during MO-GRPO training.
    We show that in the two-objective case ($K{=}2$), $Z_{\textsc{bleurt}}=Z_{\textsc{jread}}=\frac{Y}{2}$ by definition, explaining why $Y$ and $Z_i$ co-move with the same (negative) sign and an approximately $2\times$ scale difference.}
    \label{fig:mo-zy}
\end{figure}

\paragraph{Empirical reward correlations.}
Corollary~\ref{coro:MO-GRPO-std} assumes uncorrelated rewards, but BLEURT and jReadability are correlated in practice. We therefore track $Z_{\textsc{bleurt}}=Z_{\textsc{jread}}=\mathrm{Corr}(R_{\textsc{bleurt}},R_{\textsc{jread}})$ during MO-GRPO training on En–Ja (Fig.~\ref{fig:mo-zy}). For $K{=}2$, Theorem~\ref{pro:MO-GRPO-std} yields $Z_{\textsc{bleurt}}=Z_{\textsc{jread}}=Y/2$, so the figure mainly reflects the change of $Z$. Importantly, $Z$ remains far from $-1$ (and weakens over training), consistent with stable updates and reduced over-optimization to jReadability compared to GRPO. 

\paragraph{Effect of reward scale calibration.}

\begin{figure}[t]
  \centering

  \begin{subfigure}{\linewidth}
    \centering
    \includegraphics[width=0.95\linewidth]{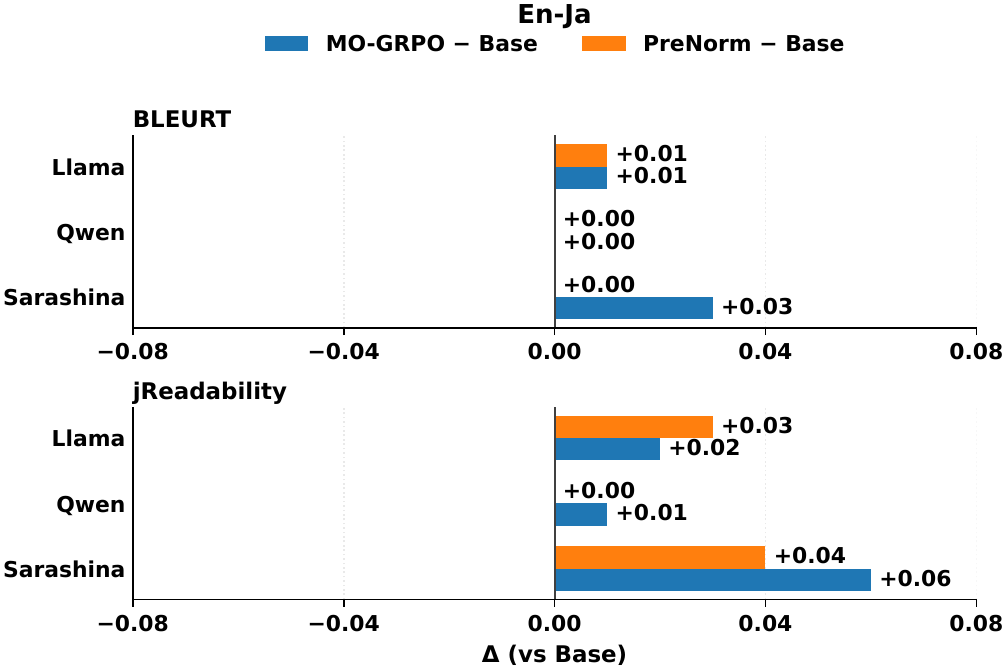}
    \caption{En-Ja dataset}
    \label{fig:delta-two-metrics-enja}
  \end{subfigure}


  \begin{subfigure}{\linewidth}
    \centering
    \includegraphics[width=0.95\linewidth]{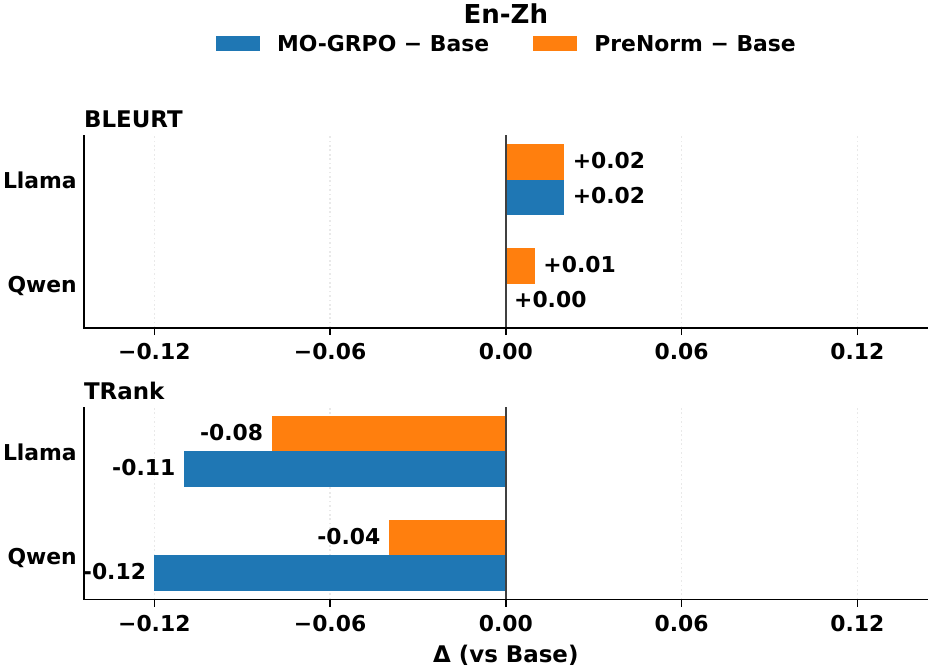}
    \caption{En-Zh dataset}
    \label{fig:delta-two-metrics-enzh}
  \end{subfigure}

  \caption{Comparison between MO-GRPO and Pre-Norm with fixed mean and standard deviation at WMT 2024. Neither exhibits reward hacking behavior, though MO-GRPO shows a slight performance improvement.}
  \label{fig:delta-two-metrics}
\end{figure}

We estimate the reward mean and variance on WMT2021 and fix them to compute advantages (Pre-Norm). With this calibration, MO-GRPO is slightly better, and neither method shows reward hacking (Fig.~\ref{fig:delta-two-metrics}). This suggests that the key is a reasonable reward scale rather than per-group normalization. Pre-Norm, however, requires an offline pass, and other WMT data (e.g., WMT2024) have different statistics among domains (Appendix~\ref{appendix:domain}), which leads to unstable learning.

\section{Related Works}
\paragraph{Multi-objective reinforcement learning.}
Multi-objective Reinforcement Learning studies sequential decision making with vector-valued rewards, where agents must trade off multiple objectives \citep{roijers2013survey}.
A common perspective seeks to approximate a set of Pareto-optimal solutions to cover multiple trade-offs \citep{roijers2013survey,hayes2022practical}.
Representative approaches explicitly maintain sets of non-dominated policies during learning \citep{JMLR:v15:vanmoffaert14a}.

\paragraph{Reward normalization in RL.}
\citet{NIPS2016_5227b6aa} normalizes value targets online to support learning across many orders of magnitude, while return-based scaling provides a simple normalization trick for improved robustness \citep{schaul2021return}.
Recent analyses show that PPO is also sensitive to reward scale and that DreamerV3-inspired implementation tricks can improve reward-scale robustness \citep{NEURIPS2023_04f61ec0}.

\paragraph{Recent variants of GRPO and relation to MO-GRPO.}
DAPO extends GRPO with asymmetric clipping, difficulty-adaptive sampling, token-level training, and explicit length control \cite{yu2025dapo}. GSPO moves from token-level to sequence-level importance ratios with sequence-level clipping \citep{zheng2025gspo}, SSPO uses subsentence-level ratios to bridge token and sequence-level updates \citep{yang2025sspo}, and collapse mitigation via entropy control \citep{simoni2025gtpo}, and improved compute efficiency via replay or pruning \citep{li2025repo,lin2025cppo}.
In contrast, MO-GRPO focuses on the objective problem, combining and normalizing multiple objectives to avoid imbalance scaling.
MO-GRPO is also applicable in many of the related studies described above.
\section{Conclusion}
We conducted an investigation into the theoretical and empirical properties of handling multiple reward functions with GRPO. Our analysis revealed a previously unreported vulnerability. The advantage function of GRPO is biased toward reward functions with high variance. This makes the algorithm susceptible to reward-hacking behaviors in multi-objective settings.
To address this weakness, we proposed Multi-Objective GRPO (MO-GRPO), an extension of GRPO that uses a simple normalization method to automatically reweight reward functions according to their value variances.
MO-GRPO treats each reward function value equitably while preserving preference orderings under rescalings.
Comprehensive experiments confirmed the practical benefits of this mechanism. We experimentally evaluate MO-GRPO in three domains: (i) the multi-armed bandits problem, (ii) machine translation tasks on the WMT benchmark (En-Ja, En-Zh), and (iii) the instruction following task. MO-GRPO consistently avoids reward hacking and shows improvements in task-specific metrics (e.g., BLEURT, jReadability) and learning stability. 

\section*{Acknowledgments}
We sincerely thank the Action Editor, Eric Yuan, and the anonymous reviewers for their insightful comments and suggestions. 
\bibliography{tacl2021}
\bibliographystyle{acl_natbib}
\clearpage
\newpage
\appendix

\section{Limitations}
There are several limitations in this paper.
(1) Our theoretical guarantees rely on asymptotic conditions, such as the number of samples approaching infinity. In practice, however, policies are updated using finite samples.
(2) Our goal is to align with human preferences, but we relied on automatic evaluators instead of conducting human evaluations.
(3) We did not validate MO-GRPO in more complex tasks, such as three or more objective problems in the real world.
(4) MO-GRPO currently combines objectives with equal coefficients after per-objective normalization.
Equal weighting is reasonable when objectives are intended to be equally important, and the main failure mode is scale mismatch.
It can be problematic when objectives reflect asymmetric priorities (e.g., safety vs.\ style) or when reward models differ in reliability.
A natural extension is weighted MO-GRPO, $A_g=\sum_i w_i \tilde A_{g,i}$, where $w_i$ encodes preferences or confidence; we leave systematic weighting strategies to future work.

\section{Reproducibility Statement}
\label{appendix:reprod}
The experiments are conducted using an NVIDIA
A100 GPU with 80 GB VRAM.

All the code of the experiments will be open-sourced upon publication. The datasets and models used in the experiments are publicly available (Table~\ref{tab:links}) except for GPT-4o-mini used for evaluation. 

\begin{table*}[t]
    
    \centering
    \small
    \adjustbox{max width=\textwidth}{
    \begin{tabularx}{\textwidth}{cX}
    \toprule
        Name & Reference \\
    \midrule
        WMT  \cite{kocmi-etal-2024-findings} &  \url{https://github.com/wmt-conference} \\\midrule
        BLEURT \cite{sellam-etal-2020-bleurt} & \url{https://huggingface.co/lucadiliello/BLEURT-20} \\\midrule
       Sarashina & \url{https://huggingface.co/sbintuitions/sarashina2.2-3b-instruct-v0.1} \\\midrule
        Qwen \cite{yang2025qwen3}& \url{https://huggingface.co/Qwen/Qwen2.5-3B-Instruct} \\\midrule
        Llama \cite{grattafiori2024llama} & \url{https://huggingface.co/meta-llama/Llama-3.2-3B-Instruct} \\\midrule
         Qwen2.5-7B-Instruct \cite{yang2025qwen3}& \url{https://huggingface.co/Qwen/Qwen2.5-7B-Instruct}\\\midrule
        Llama-3-8B-Instruct \cite{grattafiori2024llama}& \url{https://huggingface.co/meta-llama/Meta-Llama-3-8B-Instruct}\\\midrule
        jReadability \cite{hasebe2015introducing}&  \url{https://github.com/joshdavham/jreadability} \\ \midrule
        Alpaca \cite{NEURIPS2023_5fc47800}  & \url{https://huggingface.co/datasets/tatsu-lab/alpaca}\\\midrule
        RM-Mistral-7B \cite{dong2023raft} & \url{https://huggingface.co/weqweasdas/RM-Mistral-7B} \\\midrule
        TRank \cite{trokhymovych-etal-2024-open}& \url{https://huggingface.co/trokhymovych/TRank_readability}\\
        
        \bottomrule
    \end{tabularx}
    }
    \caption{List of datasets and models used in the experiments.}\label{tab:links}
\end{table*}

\section{Correlation Analysis of Dr. GRPO}
In Dr. GRPO, the advantage function is defined as 
$A^{\text{Dr}}_g=
     \sum_{i=1}^K R_i(q,o_g
)
     \;-\;
     \mathrm{mean}_\mathbf{o} \bigl(\sum_{i=1}^K R_i(q,\mathbf{o}
)\bigr).$
\begin{theorem}[Correlation each reward function and advantage function with Dr. GRPO]
Assume that  $G\rightarrow \infty$. The correlation coefficient between an individual reward function $R_i$ and the advantage $A$ is the ratio of the standard deviation of $R_i$ to the standard deviation of the total reward. 
    \begin{align*}
        &\operatorname{Corr}(R_i,A^{\text{Dr}}_g) =\\  &\frac{\sigma_i^2 + X_i}{\sqrt{\sigma_i^2\left(\sum\limits_{j=1}^K \sigma_j^2+\sum\limits_{j \neq l}\sum\limits_{l\neq j} \operatorname{Cov}\left(R_j, R_l\right)\right)}}
    \end{align*}
\end{theorem}
where $X_i = \sum_{j \neq i} \operatorname{Cov}(R_i, R_j)$.

\section{Proof of Theorem and Proposition}\label{appendix:theorem_and_pro}
From here on, for simplicity, we omit the notation for prompt $q$ and optional output $o_g$ (e.g., $R_i(q,o_g)\rightarrow R_i$). We assume the number of samples $G \to \infty$, which allows the sample statistics to approximate the true population parameters. 
\subsection{Proof of Theorem~\ref{theorem:grpo_std_un}}\label{appendix:corr}
Let $R_1, \dots, R_K$ be $K$ reward functions. Let $\mu_i = \mathbb{E}[R_i]$ and $\sigma_i^2 = \operatorname{Var}[R_i]$ denote the mean and variance of the $i$-th reward, respectively.

\begin{align*}
    &\operatorname{Cov}(R_i, A) = \operatorname{Cov}\left(R_i, \frac{ \mathbf{R} - \mathbb{E}[ \mathbf{R}]}{\sigma}\right) \\
    &= \frac{1}{\sigma} \left( \operatorname{Cov}(R_i, R_i) + \sum_{j \neq i} \operatorname{Cov}(R_i, R_j) \right) \\
    &= \frac{1}{\sigma} (\operatorname{Var}[R_i] + X_i) = \frac{\sigma_i^2 + X_i}{\sigma}
\end{align*}

Finally, we can get the correlation between each reward function and advantage function with GRPO.
\begin{align*}
\operatorname{Corr}(R_i, A) &= \frac{\operatorname{Cov}(R_i, A)}{\sqrt{\operatorname{Var}[R_i] \operatorname{Var}[A]}} \\
&= \frac{\sigma_i^2 +X_i}{\sigma \sigma_i}
\end{align*}


\subsection{Proof of Theorem~\ref{pro:MO-GRPO-std}}\label{appendix:MO-GRPO-un}
Let $R_1, \dots, R_K$ be $K$ reward functions. Let $\mu_i = \mathbb{E}[R_i]$ and $\sigma_i^2 = \operatorname{Var}[R_i]$ denote the mean and variance of the $i$-th reward, respectively.

We first calculate $\operatorname{Var}[A^{\text{MO}}]$.
\begin{align*}
\operatorname{Var}\left(A^{\mathrm{MO}}\right)&=\sum_{j=1}^K1+\sum_{l \neq j} \frac{\operatorname{Cov}\left(R_l, R_j\right)}{\sigma_l \sigma_j}\\
&=K+\underbrace{\sum_{j =1}^K \sum_{l \neq j}^K \frac{\operatorname{Cov}\left(R_l, R_j\right)}{\sigma_l \sigma_j}}_Y
\end{align*}

The corresponding correlation is:
\begin{align*}
    \operatorname{Corr}(R_i,A^{\text{MO}})&=\frac{\operatorname{Cov}(R_i,\sum_{j=1}^{K}\frac{R_j-\mu_j}{\sigma_j})}{\sigma_i\sqrt{K+Y}}\\
&=\frac{1+Z_i}{\sqrt{K+Y}}\\
\end{align*}

\subsection{Proof of Proposition~\ref{theorem:scale}}\label{appendix:scale}
    For simplicity, we know $\mu_i$ and $\sigma_i$, the true mean and standard deviation of $R_i$ over a group of outputs $\mathbf{o}$. The mean $\mu'_i$ and standard deviation $\sigma'_i$ of the transformed reward $R'_i$ are:
\begin{align*}
    \mu'_i &= \mathbb{E}[a_i R_i + b_i] = a_i \mu_i + b_i \\
    \sigma'_i &= \mathrm{std}(a_i R_i + b_i) = a_i \sigma_i \quad (\text{since } a_i > 0)
\end{align*}
The $i$-th advantage function calculated using the transformed reward $R'_i$ for any $o$ is:
\begin{align*}
   \frac{R'_i(o) - \mu'_i}{\sigma'_i}
     &= \frac{(a_i R_i(o) + b_i) - (a_i \mu_i + b_i)}{a_i \sigma_i}\\ 
     &= \frac{ R_i(o) - \mu_i}{\sigma_i}
\end{align*}
Since each advantage function is invariant, their sum $A^{\text{MO}}$ is also invariant. 

\subsection{Proof of Proposition~\ref{prop:scale}}\label{appendix:prop_scale}
For simplicity, we consider two reward functions and two outputs, $o_a$ and $o_{b}$.
Assume a trade-off scenario $R_1(o_a) > R_1(o_{b})$ and $R_2(o_a) < R_2(o_{b})$ and  $ R_1(o_{b})+ R_2(o_{b})< R_1(o_a)  +R_2(o_a) $.
We consider a scaling $\mathcal{R}'$ where $R'_i = a_i R_i$ (i.e., $b_i=0$) and derive the condition under which the preference ordering is reversed:
\begin{align*}
  &A_a > A_b\\
  &\Rightarrow A'_a < A'_b.
\end{align*}
where $A'_{a}=\frac{
     \sum_{i=1}^K R'_i(q,o_g
)
     \;-\;
     \mathrm{mean} \bigl(\sum_{i=1}^K R'_i(q,\mathbf{o}
)\bigr)}{\mathrm{std}\bigl(\sum_{i=1}^K R'_i(q,\mathbf{o}
)\bigr)}.$
This reduces to:
\begin{align*}
a_1 R_1(o_{b}) + a_2 R_2(o_{b}) &> a_1 R_1(o_a) + a_2 R_2(o_a) \\
\Rightarrow \frac{a_2}{a_1} &> \frac{R_1(o_a) - R_1(o_{b})}{R_2(o_{b
}) - R_2(o_a)}
\end{align*}
Such $a_1$ and $a_2$ exist, so GRPO does not hold.

\section{Large-Scale Models}
To verify that MO-GRPO is not specific to smaller language models, we replicate the same En-Ja training and evaluation on two larger instruction-tuned models, Qwen2.5-7B-Instruct (Qwen2.5-7B) and Llama3-8B-Instruct (Llama3-8B). We compare GRPO and MO-GRPO, and show BLEURT  and jReadability (Table~\ref{tab:Large}).
On Qwen2.5-7B, GRPO improves jReadability without degrading BLEURT, indicating no clear reward-hacking trade-off in this setting; MO-GRPO yields a small BLEURT gain while keeping readability competitive. In contrast, on Llama3-8B, GRPO increases jReadability but decreases BLEURT, suggesting a slight tendency to over-optimize jReadability at the expense of BLEURT. MO-GRPO mitigates this behavior by improving BLEURT while maintaining jReadability, demonstrating that multi-objective balancing remains important even for large-scale models.
\begin{table}[h]
  \centering
  \adjustbox{max width=\columnwidth}{
  
  \begin{tabular}{llcc}
    \toprule
     & & \multicolumn{2}{c}{\textbf{En-Ja}}\\
    \cmidrule(lr){3-4}
    \textbf{Base Model} & \textbf{Method}
      & \textbf{BLEURT}\,$\uparrow$  
      & \textbf{jReadability}\,$\uparrow$ \\
    \midrule
    \multirow{4}{*}{Qwen2.5-7B}
      & Base Model         & 0.68 & 0.66\\
      & GRPO          & 0.68 & 0.69  \\
      & MO-GRPO (ours) & 0.69 & 0.67  \\
    \midrule
    \multirow{4}{*}{Llama3-8B}
      & Base Model        & 0.59 & 0.89  \\
      & GRPO          & 0.57 & 0.92\\
      & MO-GRPO (ours) & 0.62& 0.89\\
    \bottomrule
  \end{tabular}}
  \caption{\textbf{Large-scale model results on En--Ja.} Two larger instruction-tuned LMs and compare Base Model, GRPO, and MO-GRPO. Qwen2.5-7B Instruct exhibits no clear reward-hacking trade-off, whereas Llama3-8B Instruct shows a stronger trade-off under GRPO. MO-GRPO mitigates this by recovering BLEURT while maintaining competitive readability.}

  \label{tab:Large}
\end{table}

\section{Mean and Std score per Domain in WMT-24}\label{appendix:domain}
We show per-domain score distributions on the WMT-24 datasets (\Cref{tab:ja_group_stats_compact_meanstd_2024,tab:enzh_group_stats_compact_meanstd_2024}
).
Specifically, for each model and domain, we summarize the metric values with their mean and standard deviation.
WMT-24 datasets exhibit larger domain-dependent variation about means and stds.
As a result, since mean and std vary depending on the domain, using PreNorm may cause unstable learning when training on samples from domains with different averages.

\begin{table}[h]
\small
\centering
\begin{adjustbox}{max width=\columnwidth}
\begin{tabular}{llcccccccc}
\toprule
\textbf{Metric} & \textbf{Model}
& \multicolumn{2}{c}{\textbf{Literary}}
& \multicolumn{2}{c}{\textbf{News}}
& \multicolumn{2}{c}{\textbf{Social}}
& \multicolumn{2}{c}{\textbf{Speech}} \\
\cmidrule(lr){3-4}\cmidrule(lr){5-6}\cmidrule(lr){7-8}\cmidrule(lr){9-10}
& & \textbf{Mean} & \textbf{Std}
  & \textbf{Mean} & \textbf{Std}
  & \textbf{Mean} & \textbf{Std}
  & \textbf{Mean} & \textbf{Std} \\
\midrule
\multirow{3}{*}{BLEURT}
& Llama     & 0.617 & 0.065 & 0.674 & 0.064 & 0.663 & 0.093 & 0.626 & 0.047 \\
& Qwen      & 0.636 & 0.071 & 0.672 & 0.054 & 0.684 & 0.081 & 0.625 & 0.048 \\
& Sarashina & 0.633 & 0.078 & 0.667 & 0.072 & 0.682 & 0.082 & 0.618 & 0.055 \\
\midrule
\multirow{3}{*}{jReadability}
& Llama     & 0.691 & 0.204 & 0.428 & 0.205 & 0.737 & 0.225 & 0.655 & 0.161 \\
& Qwen      & 0.678 & 0.180 & 0.441 & 0.157 & 0.705 & 0.212 & 0.661 & 0.132 \\
& Sarashina & 0.741 & 0.174 & 0.504 & 0.215 & 0.741 & 0.211 & 0.706 & 0.159 \\
\bottomrule
\end{tabular}
\end{adjustbox}
\caption{En-Ja (WMT-24): Per-domain mean and standard deviation of BLEURT and jReadability (Sample size $n$ = Literary: 206, News: 149, Social: 531, Speech: 111).}
\label{tab:ja_group_stats_compact_meanstd_2024}
\end{table}

\begin{table}[h]
\small
\centering
\begin{adjustbox}{max width=\columnwidth}
\begin{tabular}{llcccccccc}
\toprule
\textbf{Metric} & \textbf{Model}
& \multicolumn{2}{c}{\textbf{Literary}}
& \multicolumn{2}{c}{\textbf{News}}
& \multicolumn{2}{c}{\textbf{Social}}
& \multicolumn{2}{c}{\textbf{Speech}} \\
\cmidrule(lr){3-4}\cmidrule(lr){5-6}\cmidrule(lr){7-8}\cmidrule(lr){9-10}
& & \textbf{Mean} & \textbf{Std}
  & \textbf{Mean} & \textbf{Std}
  & \textbf{Mean} & \textbf{Std}
  & \textbf{Mean} & \textbf{Std} \\
\midrule
\multirow{2}{*}{BLEURT}
& Llama & 0.661 & 0.076 & 0.719 & 0.069 & 0.698 & 0.096 & 0.658 & 0.063 \\
& Qwen  & 0.712 & 0.077 & 0.760 & 0.059 & 0.747 & 0.085 & 0.695 & 0.061 \\
\midrule
\multirow{2}{*}{TRank}
& Llama & -2.946 & 1.031 & -0.661 & 1.006 & -2.651 & 1.047 & -2.855 & 1.532 \\
& Qwen  & -3.267 & 0.965 & -0.865 & 0.885 & -2.927 & 1.019 & -3.310 & 1.405 \\
\bottomrule
\end{tabular}
\end{adjustbox}
\caption{En-Zh (WMT-24): Per-domain mean and standard deviation of BLEURT and TRank (Sample size $n$ = Literary: 206, News: 149, Social: 531, Speech: 111).}
\label{tab:enzh_group_stats_compact_meanstd_2024}
\end{table}

\section{Experiment Settings in WMT}\label{appendix:parameter}
We show the parameter settings of the experiments in Table~\ref{tab:para}.
\begin{table}[b]
\small
    \centering
    \begin{tabular}{lc}
        \toprule
    Parameter&\\
        \midrule
        temperature     & 0.7  \\
        learning rate  & 2e-6 \\
        adam beta1 & 0.9\\
        adam beta2 & 0.99\\
        weight decay& 0.1\\
        gradient accumulation steps & 4\\
        num generations &8\\
        num train epochs & 3\\
         beta & 0.04\\
         LoRA rank & 128 \\
         LoRA alpha & 128\\
        \bottomrule
    \end{tabular}
    \caption{Parameter Setting of the Experiment in WMT for GRPO, Dr. GRPO, and MO-GRPO.} \label{tab:para}
\end{table}

\section{Python Implementation}\label{appendix:mo-grpo}
We show the TRL style's implementation recipe for MO-GRPO.

\lstdefinestyle{pyclean}{
  language=Python,
  basicstyle=\ttfamily\scriptsize,
  numbers=left,
numberstyle=\ttfamily\scriptsize\color{gray},
  stepnumber=0,
  numbersep=8pt,
  frame=none,
  rulecolor=\color{black!15},
  frameround=ffff,
  columns=fullflexible,
  keepspaces=true,
  showstringspaces=false,
  breaklines=true,
  breakatwhitespace=true,
  tabsize=4,
  xleftmargin=1.6em,
  framexleftmargin=1.2em,
  aboveskip=0.8em,
  belowskip=0.6em,
  keywordstyle=\color{blue!60!black},
  commentstyle=\color{green!45!black},
  stringstyle=\color{orange!60!black},
}

 \begin{lstlisting}[style=pyclean]
def MO_GRPO(rewards_per_func, num_generations, eps=1e-4):
    K = rewards_per_func.shape[-1]
    grouped = rewards_per_func.view(-1, num_generations, K)

    mean_k = nanmean(grouped, dim=1, keepdim=True)
    std_k  = nanstd(grouped,  dim=1, keepdim=True)

    # 1) per-reward advantages (MO-GRPO): [B, G, K]
    A_k = (grouped - mean_k) / (std_k + eps)
    # 2) sum across rewards -> final advantage: [B, G]
    A = nansum(A_k, dim=2)

    return A.view(-1)
\end{lstlisting}
\end{document}